\documentclass[conf]{new-aiaa}
\usepackage[utf8]{inputenc}
\hypersetup{
    colorlinks=true,
    urlcolor=blue
}

\usepackage{graphicx}
\usepackage{amsmath}
\usepackage{subcaption}
\usepackage{wrapfig}
\usepackage{bbm}

\usepackage{algorithm}
\usepackage{algorithmic}
\usepackage{multirow}

\usepackage[version=4]{mhchem}
\usepackage{siunitx}
\usepackage{longtable,tabularx}
\usepackage{soul}
\usepackage{todonotes}
\usepackage{svg}

\title{Expert Switching for Robust AAV Landing: A Dual-Detector Framework in Simulation}

\author{Humaira Tasnim\footnote{
Graduate Student, Mechanical and Nuclear Engineering Department, 
1 William L Jones Dr, Cookeville, TN 38505.}, 
Ashik E Rasul$^{*}$, 
Bruce Jo\footnote{
Associate Professor, Mechanical and Nuclear Engineering Department, 
1 William L Jones Dr, Cookeville, TN 38505.},
Hyung-Jin Yoon\footnote{
Assistant Professor, Mechanical and Nuclear Engineering Department, 
1 William L Jones Dr, Cookeville, TN 38505.}}
\affil{Tennessee Technological University, Cookeville, TN 38505 USA}

\begin{document}

\maketitle

\begin{abstract}

Reliable helipad detection is essential for Autonomous Aerial Vehicle (AAV) landing, especially under GPS-denied or visually degraded conditions. While modern detectors such as YOLOv8 offer strong baseline performance, single-model pipelines struggle to remain robust across the extreme scale transitions that occur during descent, where helipads appear small at high altitude and large near touchdown. To address this limitation, we propose a scale-adaptive dual-expert perception framework that decomposes the detection task into far-range and close-range regimes. Two YOLOv8 experts are trained on scale-specialized versions of the HelipadCat dataset, enabling one model to excel at detecting small, low-resolution helipads and the other to provide high-precision localization when the target dominates the field of view. During inference, both experts operate in parallel, and a geometric gating mechanism selects the expert whose prediction is most consistent with the AAV’s viewpoint. This adaptive routing prevents the degradation commonly observed in single-detector systems when operating across wide altitude ranges. The dual-expert perception module is evaluated in a closed-loop landing environment that integrates CARLA’s photorealistic rendering with NASA’s GUAM flight-dynamics engine. Results show substantial improvements in alignment stability, landing accuracy, and overall robustness compared to single-detector baselines. By introducing a scale-aware expert routing strategy tailored to the landing problem, this work advances resilient vision-based perception for autonomous descent and provides a foundation for future multi-expert AAV frameworks.

\end{abstract}

% =======================================================%
\section{Introduction}
% =======================================================%
Accurate and reliable helipad detection is fundamental to the safety of Autonomous Aerial Vehicles (AAVs), particularly during landing in GPS-denied or visually degraded environments~\cite{tang2023survey}. In such conditions, onboard vision often becomes the sole source of feedback for guidance and control~\cite{telikani2024machine}, and even brief perception failures can result in drift, misalignment, or catastrophic landing error. A persistent challenge arises from extreme scale variation during descent: at high altitudes the helipad appears as a small, low-resolution target, while near touchdown it expands to fill a large portion of the image~\cite{bitoun2020helipadcat}. A single detector must therefore operate across two contradictory regimes: fine-grained recognition of tiny distant objects and precise localization of large near-range structures, making robust performance fundamentally difficult.

Modern deep-learning detectors such as YOLO~\cite{redmon2017yolo9000} and SSD~\cite{liu2016ssd} have achieved strong performance in aerial-imaging benchmarks. However, these models are typically trained on datasets with limited altitude diversity and implicitly specialize toward a narrow scale range. As a result, performance often degrades when the target transitions across drastically different scales, a situation inherent in the AAV landing trajectory. Existing solutions, including detector fusion and uncertainty-based selection~\cite{gal2016dropout,lakshminarayanan2017simple,kendall2017uncertainties}, offer partial improvements but generally combine heterogeneous detectors without task-specific specialization. Meanwhile, advances in Mixture-of-Experts (MoE) architectures~\cite{masoudnia2014mixture} including DAMEX~\cite{jain2023damex}, MoE-Fusion~\cite{cao2023multi}, and CBDES~\cite{xiang2025cbdes} demonstrate that explicitly training specialized experts and routing inputs adaptively can significantly improve robustness. Yet, to our knowledge, MoE architectures have seen limited investigation for scale-adaptive perception in AAV landing, even though expert routing aligns well with the descent-scale transition.

Motivated by this gap, we propose a dual-expert YOLOv8 perception framework designed explicitly for scale-adaptive helipad detection. Two specialized detectors are trained on complementary scale regimes: a far-range expert optimized for small, low-resolution helipads, and a close-range expert optimized for large, high-resolution views. A geometric gating mechanism selects between experts based on which prediction is best aligned with the camera center, allowing the framework to adapt naturally to scale transitions during descent. Unlike classical fusion methods that rely on fixed thresholds or averaging, this hard selection strategy provides stable performance across the full altitude range while avoiding the degradation seen in single-model detectors.

The proposed framework is evaluated within a high-fidelity simulation environment that integrates photorealistic CARLA scenes~\cite{dosovitskiy2017carla} with NASA’s GUAM dynamics engine~\cite{GUAM2024}. This setup enables controlled experimentation across varied altitudes and approach paths, allowing us to assess not only detection quality but also end-to-end landing accuracy. Experimental results show improved robustness, reduced jitter in bounding-box localization, and tighter final landing error compared to single-detector baselines, highlighting the importance of scale specialization for reliable AAV landing pipelines.

In summary, the primary contributions of this work are:
\begin{itemize}
    \item \textbf{Scale-Adaptive Dual-Expert Perception Framework:} Two specialized YOLOv8 experts: far-range and near-range are trained to explicitly address the scale variability inherent in AAV landing.
    
    \item \textbf {Gating and Smooth Expert Transition:} A deterministic, scale-sensitive gating strategy combined with temporal smoothing enables stable, jitter-free perception across the descent profile.

    \item \textbf{Integrated Evaluation in Photorealistic Simulation:} The framework is validated in CARLA environments paired with NASA’s GUAM flight-dynamics model, demonstrating increased robustness and improved landing precision over single-model approaches.
\end{itemize}

\begingroup
\renewcommand\thefootnote{}
\footnote{\url{https://github.com/htasnim42/LCASL_TTU_Dual_Expert.git}}
\addtocounter{footnote}{-1}
\endgroup

\section{Literature Review}

Accurate perception during autonomous aerial vehicle (AAV) landing has attracted extensive research attention due to its safety-critical role in GPS-denied or visually degraded environments. Vision-based detection often becomes the primary sensing modality in such conditions, forming the foundation of most modern landing pipelines. However, maintaining robustness across large variations in viewing scale, environmental disturbances, and sensor noise remains a fundamental challenge in real-world autonomous landing scenarios~\cite{rasul2025development,saha2025direct}. These limitations have motivated the development of perception strategies that are not only accurate and efficient, but also explicitly adaptive to the altitude-dependent scale transitions encountered during descent.

\textbf{Object Detection for AAV Landing:} 
Deep neural networks serve as the backbone of contemporary AAV perception, with detectors such as YOLO and SSD enabling real-time inference on embedded platforms~\cite{redmon2017yolo9000,jocher2020ultralytics,liu2016ssd}. YOLO architectures, in particular, strike a practical balance between speed and accuracy and have been widely adopted in aerial imaging tasks. Several AAV-focused adaptations introduce multi-scale feature aggregation, altitude-aware anchors, and modified feature pyramids to better handle small distant helipads, as demonstrated in YOLO-DroneMS~\cite{zhao2024yolo}. Other works incorporate lightweight backbones, attention mechanisms, or multi-resolution inputs to improve precision under computational constraints.

Despite these advancements, single-model detectors still exhibit instability across the broad altitude range traversed during landing. From high elevation, helipads often occupy only a few pixels, whereas near touchdown they dominate the frame~\cite{tang2023survey}. Such drastic appearance changes challenge fixed-scale architectures, leading to inconsistent localization performance. SSD-based approaches partially mitigate this through multi-scale feature maps and anchor designs, yet abrupt scale transitions continue to degrade reliability. The HelipadCat dataset~\cite{bitoun2020helipadcat} has become a central benchmark for evaluating these behaviors, with multiple studies reporting scale variation as a primary factor behind missed detections, jitter, and false localization during descent~\cite{telikani2024machine}. These findings underscore the importance of designs that generalize across flight regimes rather than optimizing for a single altitude distribution.

\textbf{Challenges of Scale and Uncertainty in Perception:}
A key difficulty in landing perception arises from the interplay between scale variation and predictive uncertainty. Far-range helipads provide limited pixel information, increasing bounding box noise and epistemic uncertainty, while close-range targets reduce contextual cues and may produce overconfident but inaccurate predictions. This nonlinear shift in visual characteristics presents a challenge for unified detectors operating under fixed-resolution constraints.

To address reliability concerns, prior work has explored uncertainty-estimation techniques such as Monte Carlo Dropout~\cite{gal2016dropout}, deep ensembles~\cite{lakshminarayanan2017simple}, and Bayesian calibration~\cite{kendall2017uncertainties}. These approaches improve confidence modeling under distribution shift and can enhance robustness to illumination, weather, and sensor noise variations pervasive in outdoor landing scenarios~\cite{rasul2025bayesian}. Nonetheless, many existing AAV pipelines continue to rely primarily on raw detector confidence, which is often poorly calibrated during scale transitions. This gap motivates perception strategies that more explicitly account for the changing visual conditions encountered throughout descent.

\textbf{Fusion and Expert-Based Perception:}
Fusion frameworks offer another pathway toward robust perception, particularly when single-model detectors fail under challenging conditions. Multi-sensor systems often combine visual, LiDAR, radar, or inertial signals to exploit complementary strengths and reduce susceptibility to noise or occlusion~\cite{shi2020fusion}. At the model level, detector fusion across YOLO, SSD stabilize localization through complementary error patterns. However, classical fusion methods typically rely on static weighting, confidence averaging, or fixed thresholding, limiting their adaptability to rapidly evolving approach geometries.

More recent advances in Mixture-of-Experts (MoE) architectures demonstrate the potential of combining specialized experts with adaptive routing mechanisms. DAMEX~\cite{jain2023damex} employs dataset-aware experts to select the most compatible detector for each input domain, while MoE-Fusion~\cite{cao2023multi} integrates local and global experts to balance fine spatial detail with broader contextual structure, an ability particularly relevant for altitude-varying AAV imagery. CBDES MoE~\cite{xiang2025cbdes} further shows that hierarchical decoupling with attention-based routing can maintain accuracy while reducing computational overhead. Yet most MoE applications focus on heterogeneous modalities or dataset diversity rather than addressing the extreme intra-class scale variation characteristic of AAV descent. This motivates expert architectures that explicitly differentiate between far-range and near-range perception regimes and learn complementary scale biases.

\textbf{Multi-Scale Representation Learning for AAV Perception:} 
Scale variation has also driven extensive research in multi-scale representation learning. Convolutional detectors leverage spatial pyramids and multi-resolution features~\cite{he2015spatial}, while enhanced feature pyramid networks such as FPN~\cite{lin2017feature}, PANet~\cite{liu2018path}, and BiFPN~\cite{tan2020efficientdet} aim to improve cross-scale consistency. Transformer-based designs further extend this through deformable attention mechanisms~\cite{zhu2020deformable} that emphasize scale-relevant features. In aerial imagery, attention-based fusion helps preserve discriminative features across altitudes. However, these models generally employ a single unified detector, forcing one network to balance fine-detail recognition at far range with precise boundary localization at close range. This can lead to trade-offs and suboptimal performance when confronting the full descent trajectory. These limitations motivate architectures that partition scale regimes into specialized experts, an idea explored in MoE and ensemble-based methods but not systematically applied to AAV landing~\cite{masoudnia2014mixture}. The dual-expert strategy proposed in this work directly targets this gap by training separate far-range and near-range detectors and fusing them via an adaptive gating mechanism.

\textbf{Robustness to Environmental and Sensor Variations:} 
Environmental variability further complicates AAV perception. Illumination changes, shadows, motion blur, and atmospheric conditions such as fog or rain can distort helipad appearance and reduce feature reliability~\cite{bitoun2020helipadcat}. Outdoor datasets often exhibit strong distribution shift, where models trained under clear conditions degrade significantly under adverse environments. Approaches such as domain randomization~\cite{tobin2017domain}, style-transfer augmentation, and weather-adaptive training help mitigate these effects, but most still rely on a single detector attempting to generalize across all conditions~\cite{munir2024impact}. When environmental and scale-induced uncertainties interact, detection instability becomes more pronounced. Strategies that appropriately modulate prediction confidence or down-weight unreliable regimes therefore play an increasingly important role~\cite{kendall2017uncertainties}.

\textbf{Simulation-Based Evaluation in AAV Perception:} 
High-fidelity simulation platforms such as CARLA~\cite{dosovitskiy2017carla}, AirSim~\cite{shah2017airsim}, and AirTaxiSim~\cite{bansal2025airtaxisim} have become essential for evaluating landing perception systems. These environments offer controllable conditions, photorealistic rendering, and realistic sensor modeling, enabling systematic analysis of failure modes without the cost and risk of real-world experiments. Integrating simulated perception with dynamics engines such as NASA’s GUAM~\cite{GUAM2024} further captures the tight coupling between vision and control during landing. Simulation pipelines also provide repeatability across trials, supporting quantitative comparisons of detectors, fusion strategies, and decision mechanisms. These capabilities motivate the integrated CARLA-GUAM framework employed in this work.

Overall, prior work in object detection, uncertainty estimation, fusion, and multi-scale representation has established a strong foundation for AAV perception. However, major gaps remain. Single-model detectors continue to show sensitivity to scale transitions, fusion strategies often rely on static weighting, and existing MoE frameworks have not been tailored for scale adaptation within a single visual modality. These limitations motivate the scale-adaptive dual-expert perception strategy introduced in this paper.

\begin{figure}[t]
    \centering
    \includegraphics[width=0.95\linewidth]{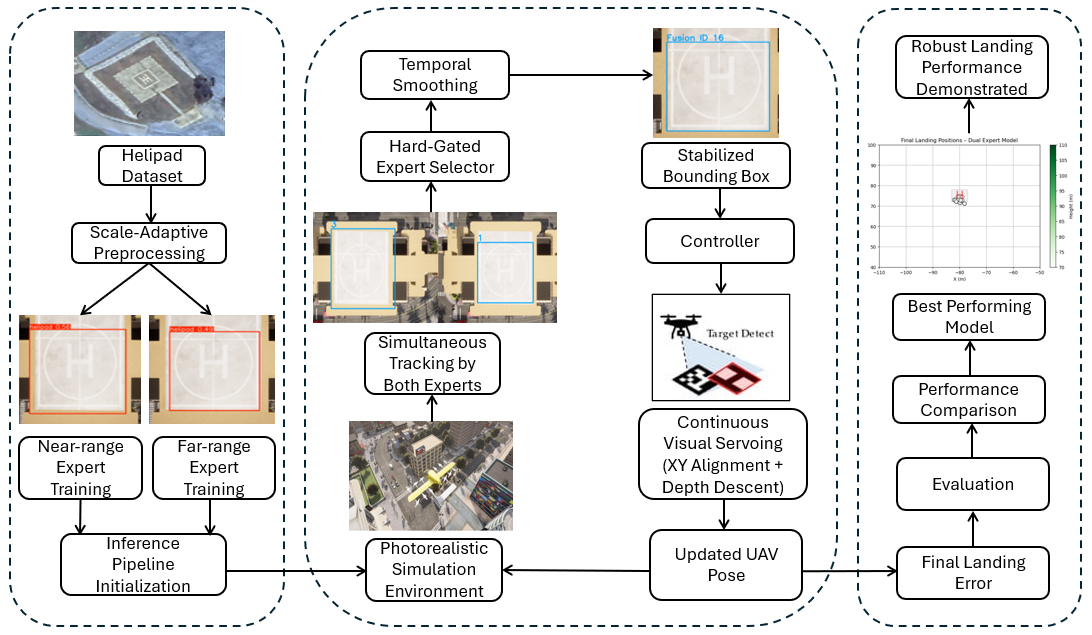}
    \caption{Overall perception-control framework integrating dataset preparation, dual-expert training, hard-gated expert selection, temporal smoothing, and visual-servoing based landing evaluation in CARLA-GUAM simulation.}
    \label{fig:overall_framework}
\end{figure}

\section{Proposed Framework}

Developing reliable perception frameworks for autonomous aerial vehicle (AAV) landing requires 
addressing visual scale variability, rapid viewpoint changes, and environmental dynamics. To overcome 
these challenges, the proposed dual-expert framework decomposes the perception task into two 
complementary scale regimes: far-range (small helipad appearance) and near-range (large helipad 
appearance) and selects between them using a deterministic hard-gating mechanism based on geometric 
alignment with the camera center. This switching strategy ensures that the perception module always 
relies on the expert best suited for the current altitude and image scale.

Figure~\ref{fig:overall_framework} summarizes the complete perception-control pipeline used in this 
work. During each descent, monocular RGB images from the downward-facing camera are streamed to two 
specialized YOLOv8 detectors operating in parallel. The far-range expert processes low-resolution 
helipad appearances at high altitudes, while the near-range expert handles close-range, large-scale 
helipad views. Their predictions are evaluated by a hard-gated expert selector that chooses the 
bounding box whose center is geometrically closest to the camera reference point. The selected box is 
then passed through a short temporal-smoothing filter to suppress frame-level jitter and produce a 
stabilized bounding-box estimate. This fused perception output drives a continuous visual-servoing 
controller that performs lateral alignment and scale-based depth descent. The controller updates the 
AAV pose within the CARLA-GUAM simulation environment, completing the closed-loop perception-control 
cycle. The final landing position is recorded for evaluation, enabling model comparison and robustness 
analysis across the ten randomized trials.

To support the dual-expert framework, each component of the pipeline requires carefully designed
training data, scale-specialized model initialization, and consistent simulation conditions. In
particular, the effectiveness of the hard-gated expert selection depends on training each detector on
a scale domain that matches its intended operating altitude. Likewise, stable closed-loop performance
relies on ensuring that the detectors generalize across the visual variations encountered throughout
descent-ranging from distant low-resolution helipad signatures to large, close-range structures.
The following subsections describe the dataset preparation, scale-adaptive preprocessing, and
training procedures used to construct the two YOLOv8 experts and integrate them into the proposed
pipeline.

\subsection{Dataset Preparation and Dual-Expert Training:}  

This work builds upon the HelipadCat dataset~\cite{bitoun2020helipadcat}, which contains approximately 4,000 annotated aerial images of helipads collected under diverse altitudes, illumination conditions, and viewpoints. Each image is labeled in YOLO format with class identifiers and normalized bounding box coordinates, enabling efficient training of modern detection architectures while preserving the inherent variability of helipad appearance across the descent trajectory.

To enable scale specialization, the dataset was processed into two distinct training sets corresponding to far-range and near-range operating regimes. All images were rescaled from their original $640{\times}640$ resolution into two configurations: $832{\times}832$ for the far-range expert, designed to detect small, low-resolution helipads at higher altitudes; and $512{\times}512$ for the near-range expert, optimized for close-range detection when the helipad occupies a substantial portion of the frame. This preprocessing strategy ensures that each expert learns features appropriate to its spatial scale domain rather than attempting to generalize across contradictory visual conditions~\cite{lin2017feature}.

Both training sets were augmented with standard transformations-horizontal flipping, random cropping, rotation, and color jitter to improve robustness while maintaining consistency across experts. The two detectors were then trained independently using the YOLOv8 architecture~\cite{jocher2020ultralytics} with identical hyperparameters (batch size of 16, 100 epochs, and SGD optimization), ensuring a fair and controlled comparison between scale regimes. YOLOv8’s decoupled detection head and spatial pyramid pooling backbone support high-resolution localization, making it suitable for both small and large object scales~\cite{he2015spatial}.

The far-range expert trained on $832{\times}832$ inputs develops sensitivity to fine-grained, low-resolution features characteristic of distant helipads, where contextual cues dominate and object detail is minimal. In contrast, the near-range expert trained on $512{\times}512$ emphasizes boundary precision, shape definition, and local texture-traits essential for reliable detection near touchdown. Validation monitoring and learning-rate scheduling were used throughout training to prevent overfitting and ensure stable convergence.

Together, the two experts exhibit complementary strengths: one excels at early detection during high-altitude approach, while the other provides reliable close-range localization during the terminal landing phase. This dual-expert foundation enables the subsequent switching mechanism to dynamically adapt perception quality across the full descent profile.

\subsection{Switching Mechanism and Expert Routing}

\begin{figure}[H]
    \centering
    \includegraphics[width=0.95\linewidth]{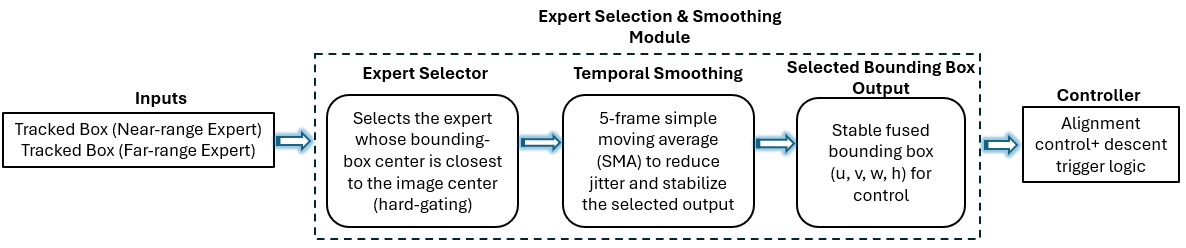}
    \caption{Expert selection module using a hard-gating strategy.}
    \label{fig:expert_module}
\end{figure}

Building on the complementary strengths of the two YOLOv8 experts, the system must determine at every frame which expert provides the most geometrically reliable helipad estimate. Since each model specializes in a different scale regime, their predictions naturally diverge across the descent trajectory. To ensure consistent, stable perception without relying on altitude measurements or confidence heuristics, we employ a lightweight expert selector mechanism demonstrated in Figure~\ref{fig:expert_module} that evaluates both experts simultaneously and selects the one whose prediction is most aligned with the camera viewpoint. Each expert produces a bounding box prediction denoted by:
\[
b_k(t) = (u_k(t), v_k(t), w_k(t), h_k(t)),
\]
where the tuple encodes the helipad location and spatial extent in pixel coordinates. Here, \(u_k(t)\) and \(v_k(t)\) represent the horizontal and vertical coordinates of the bounding-box center, while \(w_k(t)\) and \(h_k(t)\) denote the width and height of the detected helipad in the image plane. These quantities reflect how the helipad appears from the AAV’s viewpoint at time \(t\), and naturally vary with altitude, pitch, and camera motion~\cite{chaumette2006visual}. The camera’s ideal alignment reference is located at \((c_x, c_y) = (224, 224)\) for a \(448 \times 448\) input frame, assuming a standard CARLA camera model with negligible distortion~\cite{zhang2002flexible}.

Because the far-range expert is trained on upscaled images emphasizing fine-grained detail at far ranges, and the near-range expert specializes in high-resolution close-range detections, the two experts behave differently across the descent profile. Instead of averaging their predictions or selecting based on raw confidence which can be unreliable under scale changes, the system selects the expert whose bounding-box center is geometrically closest to the desired camera viewpoint~\cite{tang2021quadrotor}. This is measured using an \(L_1\)-norm distance~\cite{shazeer2017outrageously}:
\begin{equation}
    D_k(t) = |u_k(t) - c_x| + |v_k(t) - c_y|.
\end{equation}
A smaller value of \(D_k(t)\) indicates that the helipad appears closer to the image center, suggesting that the expert has provided a more reliable estimate for the current altitude.

The active expert is therefore selected using a simple hard-gating rule:
\begin{equation}
    k^\ast(t) = \arg\min_k D_k(t).
\end{equation}
At higher altitudes, the far-range expert typically produces more centered predictions because distant helipads occupy fewer pixels and better match its training distribution. As the AAV descends and the helipad grows, the near-range expert becomes more accurate and stable, resulting in a natural transition between experts without requiring explicit altitude measurements.

However, instantaneous alignment alone can lead to rapid switching near crossover altitudes where both experts produce similar detections. Such oscillations introduce jitter into the perception signal and may destabilize the downstream controller. To address this, the selected bounding box is smoothed over a short temporal window using a simple moving average:
\begin{equation}
    \hat{b}(t) = \frac{1}{N} \sum_{\tau = t-N+1}^{t} b^\ast(\tau),
\end{equation}
where \(N = 5\) in our implementation and \(b^\ast(\tau)\) is the raw selected box at time \(\tau\). This windowed averaging reduces high-frequency noise and prevents abrupt expert-switching artifacts, producing a stable fused estimate for control.

Let \((\hat{u}(t), \hat{v}(t))\) denote the center of the smoothed bounding box. The lateral image-plane errors are then defined as
\begin{equation}
    e_x(t) = c_x - \hat{u}(t), \qquad e_y(t) = c_y - \hat{v}(t),
\end{equation}
where \(e_x(t)\) and \(e_y(t)\) measure the horizontal and vertical misalignment. These errors serve as inputs to the visual-servo controller, which converts them into body-frame velocity commands~\cite{lu2018survey}. A positive \(e_x(t)\) indicates that the helipad lies to the right of the image center and the AAV must translate leftward, while negative values imply the opposite.

Since no depth sensor is used, the system infers vertical descent progress from the apparent scale of the helipad. The detected area,
\begin{equation}
    A(t) = w(t)\, h(t),
\end{equation}
increases monotonically as the AAV approaches the ground. A reference value \(A_{\mathrm{ref}}\) corresponding to the desired landing altitude is used to compute the descent error:
\begin{equation}
    e_z(t) = A_{\mathrm{ref}} - A(t).
\end{equation}
Although monocular vision cannot provide absolute depth, this scale-based measure offers a reliable proxy for range and enables stable altitude tracking.

Together, these components form a unified perception-control pipeline that delivers stable alignment, reduced noise, and robust descent behavior across challenging altitudes, including scenarios where an individual expert alone would fail.

\subsection{Simulation and Evaluation Setup:}  
The CARLA-GUAM simulation setup enables controlled yet visually diverse evaluation of the 
proposed dual-expert perception system. CARLA provides high-fidelity urban environments 
featuring realistic lighting, atmospheric variation, and complex approach geometries, while 
NASA's GUAM engine models the full six-degree-of-freedom flight dynamics of the AAV~\cite{GUAM2024}. 
Together, these systems replicate the visual and dynamical challenges encountered during 
real-world autonomous landing scenarios.

\begin{figure}[h!]
    \centering
    \includegraphics[width=\linewidth]{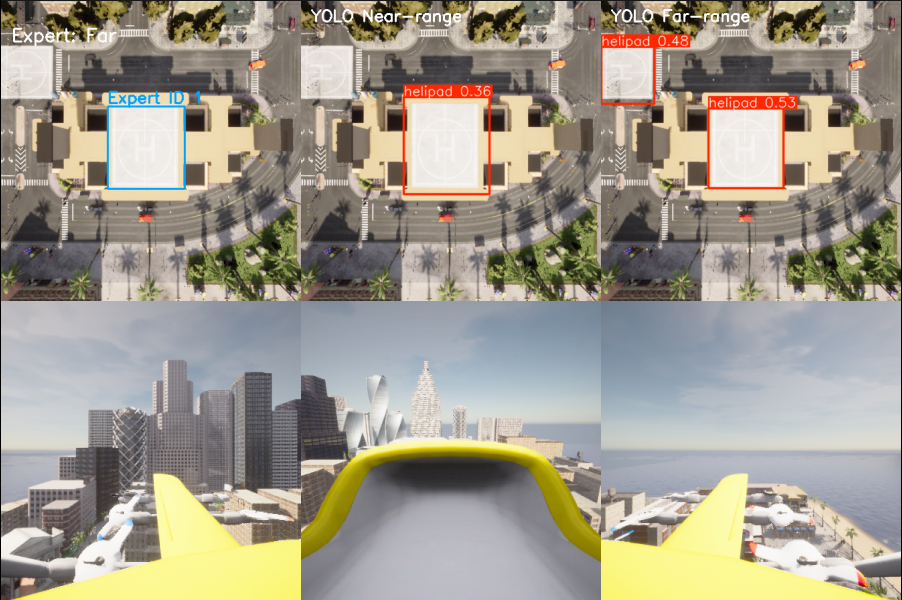}
    \caption{Real-time inference visualization during descent. 
    Top row: outputs from the Dual Expert Module, YOLO Near-range Expert, and YOLO Far-range Expert.
    Bottom row: synchronized CARLA camera views  used to validate perception consistency across viewpoints. \href{https://youtu.be/sOGckohL0J4}{https://youtu.be/sOGckohL0J4}}
    \label{fig:system_views}
\end{figure}

Figure~\ref{fig:system_views} illustrates the real-time inference pipeline during descent. 
The top row shows the outputs of the Expert Switching Module, the Near-range Expert, and the 
Far-range Expert operating in parallel. As the helipad scale changes throughout the descent, 
the two experts produce bounding boxes that reflect their respective scale specializations. 
The switching module selects the most geometry-consistent prediction at each frame and applies 
temporal smoothing to ensure stability for downstream control. The bottom row presents 
synchronized CARLA camera views, confirming that the fused detections remain consistent across 
rapid viewpoint changes.

Overall, this simulation-driven evaluation demonstrates that the proposed dual-expert framework 
maintains stable perception across wide scale variations, mitigates individual expert failure 
modes, and provides reliable bounding-box inputs for continuous visual servoing. This 
integration of scale-adaptive detection with a photorealistic simulation environment establishes 
a robust foundation for real-time autonomous landing research.

\section{Results}
\begin{figure*}[t]
    \centering
    \includegraphics[width=\textwidth]{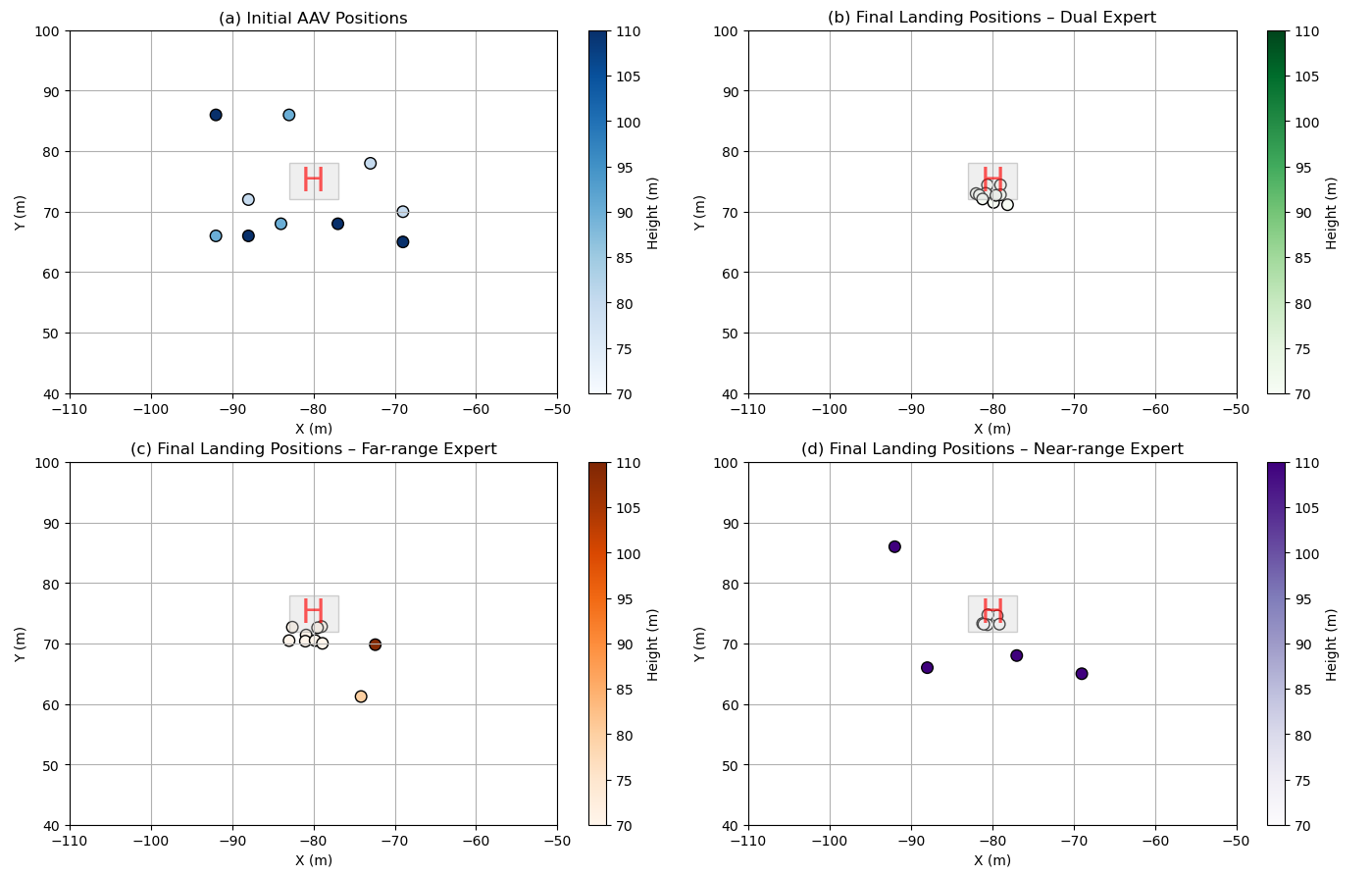}
    \caption{Landing performance comparison across all 10 trials. 
    Top-left: Initial AAV starting positions. 
    Top-right: Final landing positions using the proposed Dual Expert Model. 
    Bottom-left: Final landing positions using the Far-range Expert. 
    Bottom-right: Final landing positions using the Near-range Expert. 
    The red marker indicates the ground-truth helipad center.}
    \label{fig:landing_results}
\end{figure*}

To evaluate the effectiveness of the proposed Dual Expert Model, ten autonomous landing trials were conducted in a photorealistic CARLA-GUAM simulation environment. Each trial was initialized from a randomized $(x, y, z)$ state sampled within a controlled operating region around the helipad shown in Fig.~\ref{fig:landing_results}(a). All controllers the Near-range Expert, Far-range Expert, and the Dual Expert Model, received identical initial states and were tasked with aligning the perceived helipad bounding box with the true landing pad center while maintaining a stable descent. Figures~\ref{fig:landing_results}(b)-(d) present the final touchdown locations for each model. The red ``H'' marks the true helipad center, enabling a direct visual comparison of landing accuracy across all trials.

\subsection{Experimental Scenario and Initialization}
Randomized initial positions were sampled uniformly within a bounded region around the helipad center, located at $(x_h, y_h) = (-80, 75)\,\text{m}$.  
For each trial, the AAV’s starting position was generated using:

\[
x_0 \sim \mathcal{U}(-95,\,-65), \qquad 
y_0 \sim \mathcal{U}(60,\,90),
\]

representing a $40\text{ m} \times 30\text{ m}$ lateral window around the helipad.  
The initial altitude was sampled from the discrete set

\[
z_0 \in \{70,\; 80,\; 90,\; 110\},
\]

capturing nominal, mid-range, and long-range approach scenarios. A downward-facing RGB camera streamed real-time images to the three perception modules. No fixed flight time or descent speed was imposed; instead, the AAV descended according to the scale of the detected helipad, enabling natural closed-loop behavior driven purely by visual feedback.

\subsection{Evaluation Procedure}

For each model, the Euclidean touchdown error relative to the helipad center was computed for every
trial in which the controller successfully produced a final landing position. In cases where a model
failed to detect the helipad (e.g., the Near-range Expert in all four 110\,m trials and occasional failures of the Far-range Expert), the trial was treated as a large-error
failure case and included in the statistical analysis to reflect true operational performance. Figure~\ref{fig:landing_results} illustrates both the distribution of initial states and the 
corresponding final touchdown positions. Because the landing errors are not guaranteed to be normally 
distributed and the sample size is small ($n = 10$), a non-parametric Wilcoxon signed-rank test was used to 
compare paired touchdown errors across the models.

Table~\ref{tab:error_stats} summarizes the mean and standard deviation of final errors for the three 
controllers. The Dual-Expert model achieves the lowest average error and the smallest variability, indicating 
stable and consistent performance across diverse initial conditions.

The statistical comparison produced the following results:

\begin{itemize}
    \item \textbf{Dual-Expert vs. Far-range Expert:} Significant improvement 
    ($p = 0.0019 < 0.01$).  
    The Dual-Expert controller consistently reduced the large touchdown-error variance exhibited by the 
    Far-range Expert.

    \item \textbf{Dual-Expert vs. Near-range Expert:} No statistically significant difference 
    ($p = 0.5566 > 0.05$).  
    Although the Near-range Expert fails completely in all four 110\,m trials (no detections), its performance 
    at nominal altitudes (70--90\,m) is highly accurate, producing touchdown errors comparable to the 
    Dual-Expert model.  
    Because the Wilcoxon test evaluates only the direction of paired differences rather than their 
    magnitude, the four large failure-induced errors at 110\,m do not dominate the test statistic. This confirms 
    that the Dual-Expert model preserves the strong nominal-altitude accuracy of the Near-range Expert while 
    eliminating its long-range failure modes.
\end{itemize}

These results indicate that the Dual Expert Model maintains the high accuracy of the Near-range 
Expert under nominal altitudes while providing a substantial robustness advantage over the Far-range Expert, 
successfully completing all trials without catastrophic failures.
\begin{table}[H]
\centering
\caption{Mean and standard deviation of touchdown errors across all ten trials.}
\begin{tabular}{lcc}
\hline
\textbf{Model} & \textbf{Mean Error (m)} & \textbf{Std Dev (m)} \\
\hline
Dual-Expert & 2.53 & 1.03 \\
Near-range Expert & 5.53 & 5.57 \\
Far-range Expert & 5.60 & 3.82 \\
\hline
\end{tabular}
\label{tab:error_stats}
\end{table}

\subsection{Landing Accuracy Across Nominal Altitudes (70-90) m}
Across the ten randomized trials, the Near-range Expert achieved the lowest mean landing error under nominal altitudes. Because this model is trained on large-scale, close-range visual patterns, it produces stable and well-centered bounding-box estimates during the lower portions of the descent. However, the Near-range Expert demonstrated mild sensitivity to scale transitions, and several trials exhibited small deviations during the final meters of landing as the helipad grew rapidly in the image plane. The Far-range Expert successfully detected the helipad at all nominal altitudes but exhibited the largest variance in final touchdown positions. Since this model is optimized for distant, small-scale visual appearances, its bounding-box predictions tend to fluctuate more at close range, where subtle pixel-level noise is amplified. These fluctuations accumulate into positional drift during terminal descent, resulting in less consistent landing accuracy.

The Dual Expert Model achieved competitive accuracy relative to the Near-range Expert while maintaining substantially lower variance across trials. By dynamically selecting the expert whose bounding-box center was closest to the camera center, the switching model reduced jitter, stabilized alignment, and preserved consistent descent behavior throughout the approach.

\subsection{High-Altitude Robustness (110 m Trial)}

The 110\,m initialization represents the most challenging scenario because the helipad appears extremely 
small at the beginning of descent. At this altitude, the Near-range Expert failed across all trials, producing 
no usable detections and resulting in immediate tracking loss. This behavior is expected, as the model is 
trained for near-range, large-scale helipad appearances and cannot generalize to the extremely small visual 
footprint present at 110\,m.

The Far-range Expert successfully completed most of the 110\,m trials. One trial, however, exhibited a 
detection failure caused by severe scale compression and scene-level ambiguity within the CARLA environment. 
Two rooftop structures displayed helipad-like features at nearly identical scales, causing the detector to 
briefly lock onto the incorrect target before losing track. This reflects a known limitation of single-expert, 
single-scale vision systems operating in visually repetitive environments rather than a deficiency of the 
model itself.

The proposed Dual Expert Model successfully completed all 110\,m trials. Its adaptive expert routing 
and temporal smoothing mitigated the frame-to-frame jitter affecting the Far-range Expert and preserved a 
stable bounding-box signal throughout the descent. This experiment highlights a key limitation of fixed 
single-expert controllers: even when one expert is well-suited for far-range perception, extreme scale 
compression and scene ambiguity can cause failures. The switching mechanism provides robustness by 
dynamically compensating for such weaknesses.

\subsection{Overall Performance Comparison}

Across all trials:

\begin{itemize}
    \item \textbf{Near-range Expert}: Most accurate under nominal altitudes but fails entirely in the long-range ($110\,\text{m}$) case.  
    \item \textbf{Far-range Expert}: Robust to all altitudes but exhibits the largest touchdown variability.  
    \item \textbf{Expert Switching Model}: Completes all ten trials successfully and offers the best trade-off between accuracy, variance reduction, and robustness.
\end{itemize}

Overall, the proposed Expert Switching Model achieves a 100\% mission-success rate while maintaining competitive landing accuracy.  
Its ability to transition between experts based on geometric alignment significantly improves resilience to failure modes and mitigates large alignment errors during the final stages of descent.

\section{Conclusions and Future Work}

The proposed dual-expert perception framework demonstrates that explicitly separating far-range and close-range detection regimes significantly improves the robustness of visually guided AAV landing. By training two scale-specialized YOLOv8 detectors and integrating them through a simple yet effective expert-switching mechanism, the system maintains stable helipad localization across large altitude changes a challenge that single detectors often fail to address reliably. Simulation experiments conducted within the integrated CARLA-GUAM environment showed that the switching strategy not only improves detection stability but also yields more consistent landing trajectories, particularly in high-altitude scenarios where the near-range expert fails and the far-range model alone becomes noisy. The success of the switching-based pipeline, even under multi-helipad interference and scale-induced jitter, highlights the importance of scale-aware specialization for real-time descent operations.

Although the switching logic employed in this work is intentionally lightweight, it opens several avenues for deeper investigation. A key direction for future research is the development of a reinforcement-learning-based gating mechanism capable of learning expert selection policies directly from landing performance rather than relying on handcrafted geometric rules. Such an adaptive policy could reason over temporal context, uncertainty, and scene dynamics, enabling smoother transitions and more intelligent expert usage throughout the descent. In parallel, incorporating uncertainty estimation into the perception pipeline for example, via ensemble-based variance, Bayesian approximations, or learned confidence measures would allow the controller to account for fluctuating reliability and avoid actions driven by unstable detections. Finally, while simulation offers controlled and repeatable experimentation, extending the framework to real-world AAV testing will be essential for validating scalability under natural lighting, sensor noise, platform vibration, and environmental disturbances. Together, these directions move toward a more adaptive, uncertainty-aware, and fully learned perception-control architecture for autonomous landing.

\bibliography{mybib}

\end{document}